\documentclass{article}

     \PassOptionsToPackage{numbers, compress}{natbib}
     \usepackage[final]{neurips_2023_ml4ps}

\usepackage[utf8]{inputenc} % allow utf-8 input
\usepackage[T1]{fontenc}    % use 8-bit T1 fonts
\usepackage{hyperref}       % hyperlinks
\usepackage{url}            % simple URL typesetting
\usepackage{booktabs}       % professional-quality tables
\usepackage{amsfonts}       % blackboard math symbols
\usepackage{nicefrac}       % compact symbols for 1/2, etc.
\usepackage{microtype}      % microtypography
\usepackage{xcolor}         % colors

\usepackage{bm}
\usepackage{amsmath}
\usepackage{amsfonts}
\usepackage{amssymb}
\usepackage{mathtools}
\usepackage{mathdots}
\usepackage{physics}

\usepackage{listings}

\title{Fast Detection of Phase Transitions\\
with Multi-Task Learning-by-Confusion}

\author{%
 Julian Arnold\\
  Department of Physics\\
  University of Basel\\
  Klingelbergstrasse 82, 4056 Basel, Switzerland \\
  \texttt{julian.arnold@unibas.ch}
   \And
   Frank Schäfer\\
   CSAIL\\
  Massachusetts Institute of Technology\\
  Cambridge, Massachusetts 02139, USA\\
  \texttt{franksch@mit.edu}
  \And
    Niels Lörch\\
  Department of Physics\\
  University of Basel\\
  Klingelbergstrasse 82, 4056 Basel, Switzerland \\
  \texttt{niels.loerch@unibas.ch} \\
  }

\begin{document}

\maketitle

\begin{abstract}
Machine learning has been successfully used to study phase transitions. One of the most popular approaches to identifying critical points from data without prior knowledge of the underlying phases is the \emph{learning-by-confusion} scheme. As input, it requires system samples drawn from a grid of the parameter whose change is associated with potential phase transitions. Up to now, the scheme required training a distinct binary classifier for each possible splitting of the grid into two sides, resulting in a computational cost that scales linearly with the number of grid points. In this work, we propose and showcase an alternative implementation that only requires the training of a \emph{single} multi-class classifier. Ideally, such multi-task learning eliminates the scaling with respect to the number of grid points. 
In applications to the Ising model and an image dataset generated with Stable Diffusion, we find significant speedups that closely correspond to the ideal case, with only minor deviations.

\end{abstract}

\section{Introduction}

An exciting application of machine learning in physics is the
detection of phase transitions from data~\cite{wang:2016,carrasquilla:2017,van:2017,wetzel1:2017,carleo:2019,schaefer:2019,greplova:2020,carrasquilla:2020,arnold:2021,carrasquilla:2021,dawid:2022}: the state of a physical system is sampled at different values of a tuning parameter that determines the distribution of samples. The states are fed to an algorithm that identifies the parameter values at which the system undergoes a phase transition. 

One of the most popular methods to accomplish this task is \emph{learning-by-confusion}~\cite{van:2017}, which has successfully revealed phase transitions in a large variety of physical systems using data from both simulation~\cite{van:2017,liu:2018,beach:2018,suchsland:2018,lee:2019,guo:2020,kharkov:2020,corte:2021,bohrdt:2021,arnold:2022,richter:2022,gavreev:2022,zvyagintseva:2022,schlomer:2023,guo:2023,arnold:2023} and experiment~\cite{bohrdt:2021}. 
So far, its implementation has been computationally costly, as it involves training $K$ distinct binary classifiers and analyzing their accuracy, where $K+1$ is the number of sampled values of the tuning parameter. 

In this work, we propose an alternative implementation of the learning-by-confusion scheme that works by training a \emph{single} $K$-class classifier, which thus promises a speedup by a factor of $K$ in the ideal case. We demonstrate a significant speedup in two applications: First, the thermal phase transition of the Ising model, which is theoretically well-understood and allows for a comparison with established results. Second, an image dataset generated with Stable Diffusion~\cite{rombach2021highresolution}, which serves as a more challenging example where no prior knowledge of transitions is available.

\section{Unbiased Learning-by-Confusion with Multi-Task Learning}
\begin{figure*}[tbh]
	\centering
		\includegraphics[width=1\linewidth]{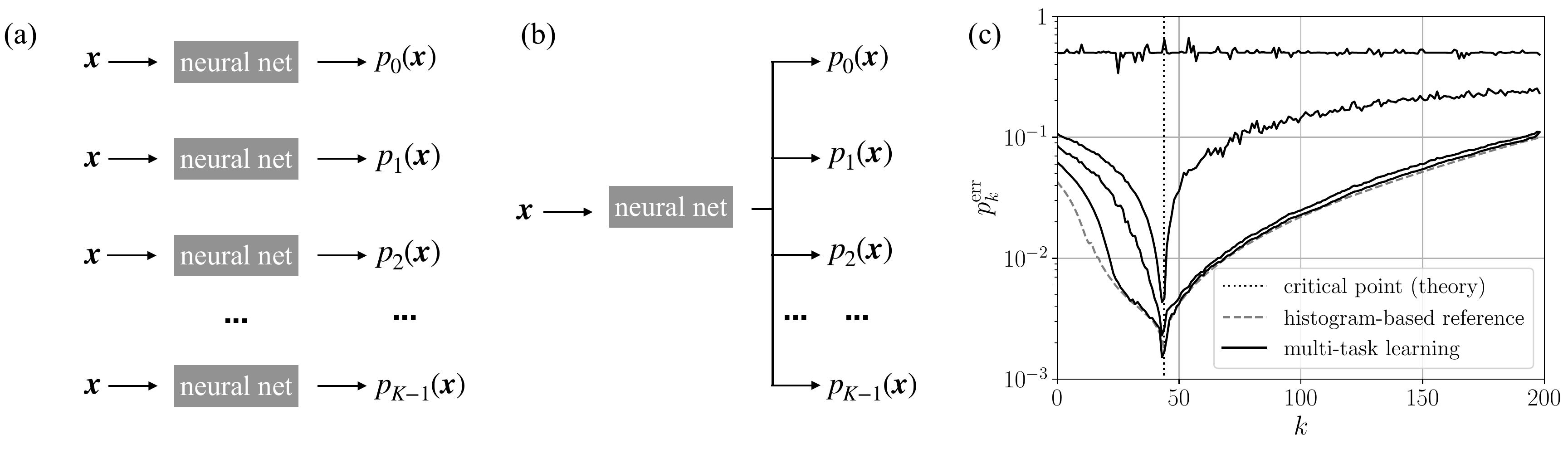}

		\caption{Schematic illustration of the learning-by-confusion method for detecting phase transitions (a) with original single-task architecture and (b) with our proposed multi-task architecture.
  (c) Classification error for the Ising model at each node using a single network trained on spin configurations. The solid lines correspond to, from top to bottom, the result after 0, 1, 5, and 50 epochs of training averaged over 5 independent runs. The vertical dotted line indicates the true location of the phase transition and the dashed line corresponds to an estimate of the Bayes-optimal error rate obtained using a histogram-based generative classifier in energy space~\cite{arnold:2022,arnold:2023}.
  }
		\label{fig:1}
\end{figure*}

In the simplest use case of the learning-by-confusion method~\cite{van:2017}, a physical system undergoes a phase transition as a function of a single real-valued parameter $\theta$. To detect the critical point $\theta^*$ at which the phase transition occurs, the $\theta$-axis is discretized into $K+1$ different points and at each point, $M$ samples are drawn from the system. With $K+1$ points, there are $K$ possibilities to separate the $\theta$-axis in two non-empty, contiguous regions $\Theta_{k}^{<} = \{ \theta|\theta < \theta^*_k\}$ and $\Theta_{k}^{>} = \{ \theta| \theta > \theta^*_k \}$, each corresponding to a tentative location $\theta^*_k = (\theta_k + \theta_{k+1})/2$ of the phase transition that lies between the grid points $\theta_k$ and $\theta_{k+1}$.
For each of these splittings, a separate classifier is trained to distinguish the two corresponding classes of samples, see Fig.~\ref{fig:1}(a). Intuitively, whichever classifier $k$ is least confused, i.e., achieves the lowest error rate on evaluation, must have been trained on the most natural splitting of the data. Therefore, the value $\theta^*_k$ associated with the lowest error rate is our best guess for the location of the phase transition.

The loss function of the $k$th classifier is an unbiased binary cross-entropy loss
\begin{equation}
\label{loss}
    \mathcal{L}_k = - \frac 1 {2} \sum_{y \in \{ >, < \}} \frac 1 { |\mathcal{D}^{y}_k|}  \sum_{\bm{x} \in \mathcal{D}^{y}_k} {\log \left[\hat p_k(y|\bm{x})\right]} ,
\end{equation}
where $|\mathcal{D}^{y}_k|$ is the size of the dataset $\mathcal{D}^{y}_k$ of samples $\bm{x}$ corresponding to ${\Theta}^{y}_k$, and $\hat p_k$ is the estimated class probability of the classifier. The normalization $\frac 1 {|\mathcal{D}^{y}|}$ compensates for the imbalance of classes that would otherwise bias the signal and may lead to a failure of the confusion scheme~\cite{arnold:2023}. Similarly, the error rate of the $k$th classifier can be estimated as
\begin{equation}
\label{perr}
    p^{\rm err}_k \approx \frac 1 {2} \sum_{y \in \{>,< \}} \frac{1}{|\mathcal{D}^{y}_k|}\sum_{\bm{x} \in \mathcal{D}^{y}_k} \mathrm{err}_k(y, \bm x),
\end{equation}
where for each sample $\bm x$ the error $\mathrm{err}_k(y, \bm x)$ is 0 if it is classified correctly and 1 if it is classified erroneously.
During training, $\mathcal{D}^{y}_k$ in Eq.~\eqref{loss} refers to a training set, while it typically refers to a separate evaluation set when estimating the error via Eq.~\eqref{perr}.

Instead of training a new classifier for each tentative splitting $k$, we propose to train a single classifier with $K$ outputs $\{ \hat p_k\}_{k=0}^{K-1}$, cf. Figs.~\ref{fig:1}(a) and (b). The loss function for this multi-task architecture is $\mathcal{L} = \frac 1K \sum_{k=0}^{K-1} \mathcal{L}_k$.
 The evaluation of the error rate remains the same as before.
 Multi-task learning~\cite{caruana:1997,ruder:2017} is expected to be highly efficient because the $K$ classification tasks are very similar to each other and only differ in a slight alteration of the tentative splitting of the parameter space. Thus, the learned features are very much transferable between tasks.

\section{Benchmark and Application}
\label{sec:applications}

\subsection{Ising Model}

\indent As a test system that has been extensively analyzed theoretically and for which established benchmarks are available, we consider the two-dimensional square-lattice ferromagnetic Ising model. It is described by the energy function
    $E(\bm{\sigma}) = - J \sum_{\langle ij\rangle} \sigma_{i}\sigma_{j},$
where the sum runs over all nearest-neighboring sites (with periodic boundary conditions), $J$ is the interaction strength $(J>0)$, and $\sigma_{i} \in \{+1,-1 \}$ denotes the discrete spin variable at lattice site $i$. 
The Ising model exhibits a phase transition between a paramagnetic 
phase at high temperature $T$ and a ferromagnetic
phase at low temperature \cite{onsager:1944}.

To generate the dataset for the Ising model, we sample spin configurations $\bm{\sigma}$ on a $60 \times 60$ lattice from Boltzmann distributions at 200 equally-spaced parameter values between $0.05$ $k_{\rm B}T/J$ and $10$ $k_{\rm B}T/J$ via Markov chain Monte Carlo, see Appendix~\ref{app:ising} for details. Figure~\ref{fig:1}(c) shows how the learning-by-confusion signal of a multi-task convolutional neural network trained on this dataset evolves with training epochs, see Appendix~\ref{app:ising} for implementation details. Eventually, the node achieving the lowest error rate coincides with the critical point.

\begin{figure}[tbh!]
	\centering
  		\includegraphics[height=0.35\linewidth]{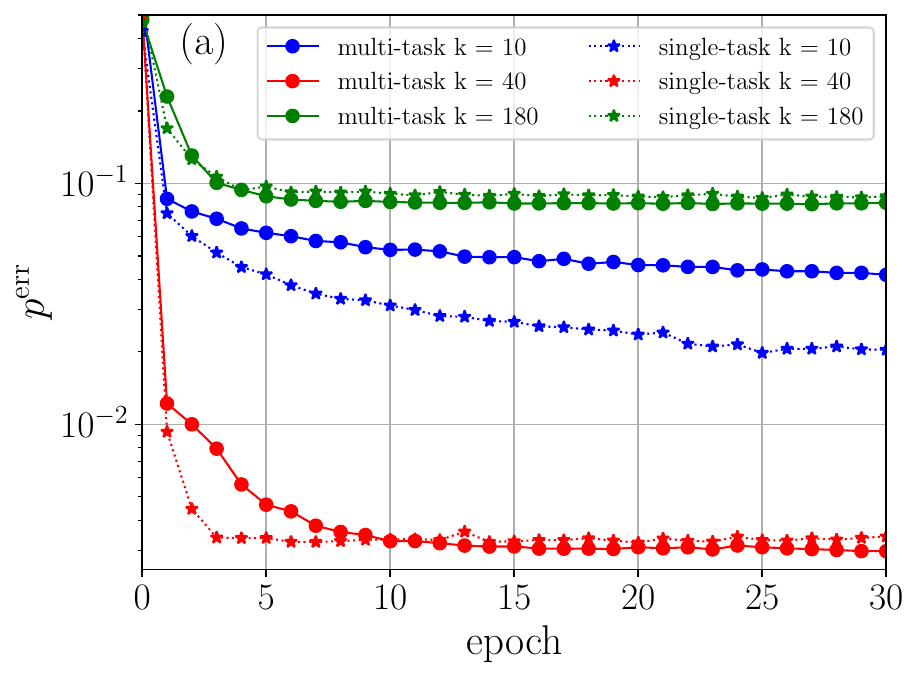}		\includegraphics[height=0.35\linewidth]{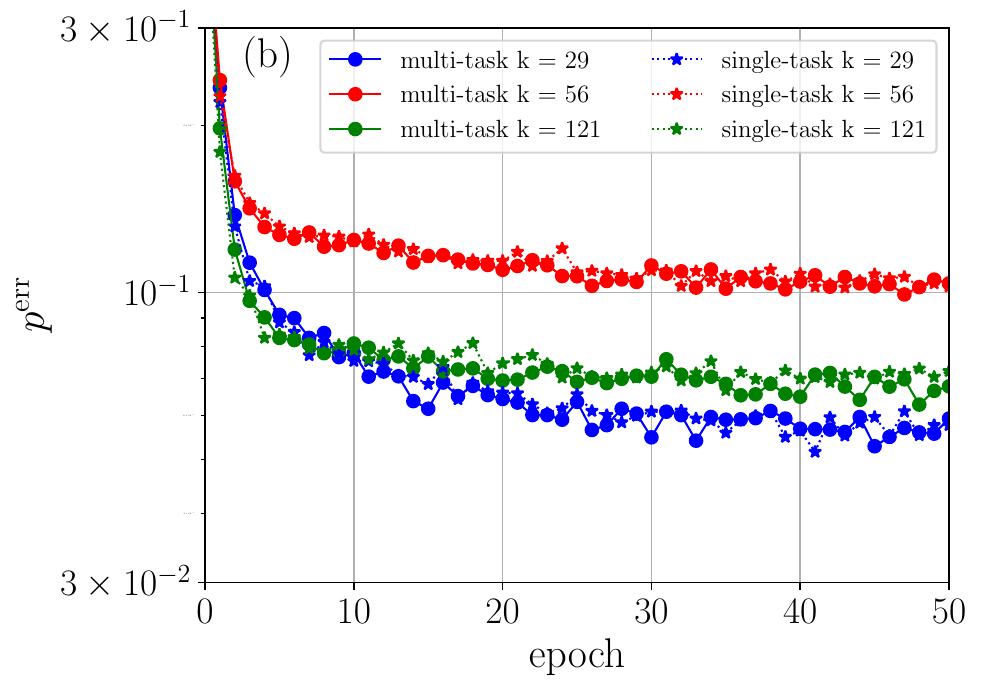}
		\caption{Error rate at representative nodes as a function of training epoch for single-task and multi-task learning-by-confusion for (a) the Ising dataset averaged over 5 runs and (b) the Stable Diffusion dataset averaged over 4 runs. Error bars derived from the standard deviation are negligible (same scale as markers), and underestimate the true confidence intervals, e.g., due to the non-Gaussian nature of the data.
  }
		\label{fig:2}
\end{figure}

To compare the single- and multi-task approach, we train 4-layer convolutional networks with otherwise identical architectures and training settings. Figure~\ref{fig:2}(a) shows how the error rate evolves as a function of the training epoch at nodes below, near, and above the critical point. In all cases, single-task learning-by-confusion shows a slightly faster rate of convergence early on during training. Below the phase transition, multi-task learning-by-confusion does not achieve an error rate as low as single-task learning-by-confusion. In contrast, near and above the phase transition, multi-task learning-by-confusion eventually catches up and even achieves lower error rates.

We also studied the training behavior of multi-task networks that each have an output node corresponding to the true critical point as well as a varying number of additional output nodes. In particular, we recorded the number of epochs it takes to reach different error thresholds at their critical output node. At low thresholds, the difference in the number of epochs between different architectures was negligible. At high thresholds, we observed the number of epochs to marginally increase with the number of output nodes. However, we observed no clear scaling and any overhead was minor (at worst $\approx \times  6$ for some combinations of training hyperparameters and model architectures) as compared to the speedup gained through multi-tasking.

\subsection{Stable Diffusion}\label{sec:stable_diff}

We now consider an image dataset generated using Stable Diffusion~\cite{rombach2021highresolution}, where for each integer $\theta$ in $[1900, 2050)$ images are sampled with the prompt ``technology of the year $\theta$''. For example, node 0 corresponds to the separator between the years 1900 and 1901. While this dataset does not feature phase transitions in the physical sense, the learning-by-confusion scheme can be used to identify points in parameter space at which the data distribution changes rapidly. Note that no prior benchmark is available for this data and the predicted change points cannot be verified by theory. As the images have complex features comparable to typical image datasets, we load the pretrained ResNet-50~\cite{He_2016_CVPR} from \texttt{PyTorch}~\cite{paszke:2019}, and exchange its final layer to fit our task, see Appendix~\ref{app:stable} for implementation details.

\begin{figure}[h!]
	\centering
  		\includegraphics[width=0.8\linewidth]{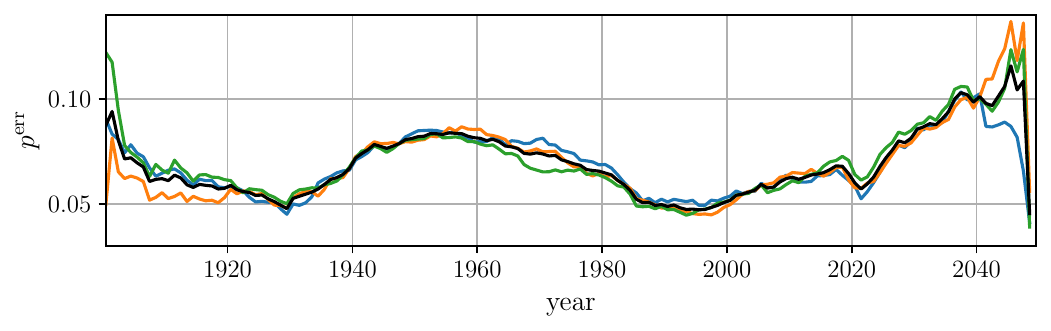}	
		
		\caption{Error rate obtained using multi-task learning-by-confusion for the Stable Diffusion dataset as a function of $\theta$ in prompt "technology of the year $\theta$". The colored lines depict the results obtained by training on three separate datasets, each averaged over 10 runs as described in Appendix~\ref{app:implementation} in more detail. The black line represents their mean.
  }
		\label{fig:3}
\end{figure}

Figure~\ref{fig:3} shows (at least) three major local minima indicating rapid changes in the image dataset; one between the years 1929 and 1930, a second, broader one in the 1990s, and a third one between 2021 and 2022. The Stable Diffusion model has only encountered images of actual technology from before and around 2022 in its training dataset LAION-5B~\cite{schuhmann2022laion}, which may explain the third minimum. Due to the small dataset size and the resulting challenges in generalization, the signal is generally less reliable close to the edges.

Figure~\ref{fig:2}(b) shows the error at representative points, the extrema at nodes 29, 56, 121 (corresponding to the transitions between years 1929-1930, 1956-1957, and 2021-2022) as a function of the training epoch for a single-task and multi-task network with otherwise identical network architecture and training parameters.
As there is no significant overhead, the speedup of the multi-task approach is approximately given by the number of grid separators (here $K=149$).

\section{Discussion and Conclusion}
\label{sec:discussion}
In the limit of infinite model capacity, both multi-task and single-task learning models ultimately yield the same predictions, because the multi-task loss corresponds to the average of the single-task losses. However, in real-world scenarios where model expressivity, training time, and data are limited, the learning behavior depends on the particulars of the model and dataset at hand.

For the Ising dataset analyzed with a shallow 4-layer convolutional net, we observed some differences in predictions between the single-task and multi-task architectures, but not near the critical point where it would matter most. At the critical point, we found a minor overhead with respect to the ideal speedup linear in the number of grid points.

The analysis of the Stable Diffusion dataset with the 50-layer ResNet-50 demonstrates the viability of the multi-task learning-by-confusion algorithm to reveal rapid changes in the distribution of a complex dataset, where a theoretical description is not available and a larger model is required to learn the features. In this case, we found no signs of an overhead.
This is in line with our general expectation that the relative overhead associated with the multi-task approach decreases as the underlying classification tasks get more complicated.

In conclusion, we find the multi-task implementation of the learning-by-confusion algorithm to provide much faster execution on large parameter grids as compared to its single-task version.

\section*{Broader Impact}
The characterization of phases of matter and the study of critical phenomena are of great importance in physics. Our work contributes a faster variant of a highly popular unsupervised learning method for the data-driven detection of phase transitions. 

By revealing structure in the output images of the Stable Diffusion generative model, we demonstrate an application of the learning-by-confusion method for change point detection beyond physics. For datasets outside statistical physics, the demonstrated speedup of our multi-task approach is particularly impactful as their analysis typically requires a large amount of computational resources.

\section*{Acknowledgments and Disclosure of Funding}
We thank Christoph Bruder for stimulating discussions and helpful suggestions on the manuscript. J.A. and N.L. acknowledge financial support from the Swiss National Science Foundation individual grant (grant no. 200020 200481). This material is based upon work supported by the National Science Foundation under grant no. OAC-1835443, grant no. OAC-2103804, and grant no. DMS-2325184.

\bibliographystyle{apsrev4-1}
\bibliography{refs.bib}

%%%%%%%%%%%%%%%%%%%%%%%%%%%%%%%%%%%%%%%%%%%%%%%%%%%%%%%%%%%%

\appendix
\section{Appendix}
\label{sec:appendix}

\subsection{Implementation}\label{app:implementation}
A \texttt{Python} implementation of the single-task and multi-task learning by confusion method is available at \cite{multitaskLBC2023}.
The hyperparameters used to generate the figures in this article are summarized in Table~\ref{tab:parameters}. Here, $M_{\text{train}}$ and $M_{\text{valid}}$ refer to the number of samples per grid point within the training and validation set, respectively. For training, we use the Adam optimizer~\cite{kingma:2014} with a learning rate given in Table~\ref{tab:parameters}.

\begin{table}[h]
\centering
\caption{Hyperparameters employed in this paper (default settings are used except where explicitly stated).}
\begin{tabular}{|c|c|c|c|c|c|}
\hline
Figure & $M_{\text{train}}$ & $M_{\text{valid}}$ & batch size & learning rate & training epochs\\
\hline
\ref{fig:1}(c) & 1000 & 1000 & 1024 & $1 \times 10^{-4}$ & 50 \\
\hline
\ref{fig:2}(a) & 1000 & 1000 & 1024 & $1 \times 10^{-4}$ & 30 \\
\hline
\ref{fig:2}(b) & 1000 & 1000 & 1024 & $5 \times 10^{-4}$ & 50 \\
\hline
\ref{fig:3} & 3 $\times$ 16 & 3 $\times$ 11 & 256 & $5 \times 10^{-5}$ & 150 \\
\hline
\end{tabular}
\label{tab:parameters}
\end{table}

\subsubsection{Stable Diffusion}
\label{app:stable}
The total Stable Diffusion dataset contains $3 \times 27$ images per year. In Fig.~\ref{fig:3}, the assignment of training and validation set is randomized before each run, so the distinction between training set and validation set does not strictly apply. The number of training epochs was set to 150 as the overall validation loss started increasing again around that point. Because the accuracy at individual nodes may have peaked earlier already, each curve represents the minimal error across all epochs.
In Fig.~\ref{fig:2}, we only use one of the three subsets making up the Stable Diffusion dataset corresponding to the seed numbers $N < 27$ in the accompanying \texttt{Python} script, and the images with seed numbers $N<16$ are fixed as the training set.
A \texttt{Python} script with seeds to generate the dataset and a notebook to perform training can be found at \cite{multitaskLBC2023}.
\subsubsection{Ising Model}
\label{app:ising}

The Ising dataset is generated by sampling spin configurations from Boltzmann distributions at various temperatures via the Metropolis-Hastings algorithm. The lattice is initialized in a state with all spins pointing up and updated by drawing a random spin that is flipped with probability ${\rm min}(1, e^{-\Delta E/k_{\rm B} T})$, where $\Delta E$ is the energy difference resulting from the spin flip. In a thermalization period, we sweep the complete lattice $10^5$ times. Afterward, we collect samples, increase the temperature, and start another thermalization period.

For the Ising model, we utilized fixed training and validation sets, as randomization in the split was unnecessary due to the abundance of data.
The convolutional neural network architecture utilized to produce the results in Figs.~\ref{fig:1}(c) and~\ref{fig:2}(a) can be found in the accompanying code \cite{multitaskLBC2023}.

\subsection{Compute Resources}
\label{app:compute_resources}
For our computations, we use an NVIDIA GTX 3090 GPU and an Intel i9-10900K CPU, where a training and validation epoch finishes within a few seconds for both the Ising and Stable Diffusion dataset.

\end{document}